\documentclass[10pt,twocolumn,letterpaper]{article}

\usepackage{cvpr}
\usepackage{times}
\usepackage{epsfig}
\usepackage{graphicx}
\usepackage{amsmath}
\usepackage{amssymb}

\usepackage{subcaption}

\usepackage{authblk}

\usepackage[breaklinks=true,bookmarks=false]{hyperref}

\cvprfinalcopy % *** Uncomment this line for the final submission

\def\cvprPaperID{****} % *** Enter the CVPR Paper ID here

\begin{document}

%%%%%%%%% TITLE
\title{Introduction to the 1st Place Winning Model of OpenImages Relationship Detection Challenge}

% \author{Ji Zhang\\
% {\tt\small zhangjiapply@gmail.com}

\renewcommand{\thefootnote}{\fnsymbol{footnote}}
\renewcommand\Authands{, }
\makeatletter
\renewcommand\AB@affilsepx{, \protect\Affilfont}
\makeatother

\author[1,2]{Ji Zhang}
\author[1]{Kevin Shih}
\author[1]{Andrew Tao}
\author[1]{Bryan Catanzaro}
\author[2]{Ahmed Elgammal}
\affil[1]{Nvidia Research}
\affil[2]{Rutgers University} 

\maketitle
%\thispagestyle{empty}

%%%%%%%%% ABSTRACT
\begin{abstract}
   This article describes the model we built that achieved 1st place in the OpenImage Visual Relationship Detection Challenge on Kaggle. Three key factors contribute the most to our success: 1) language bias is a powerful baseline for this task. We build the empirical distribution $P(predicate|subject,object)$ in the training set and directly use that in testing. This baseline achieved the 2nd place when submitted; 2) spatial features are as important as visual features, especially for spatial relationships such as ``under'' and ``inside of''; 3) It is a very effective way to fuse different features by first building separate modules for each of them, then adding their output logits before the final softmax layer. We show in ablation study that each factor can improve the performance to a non-trivial extent, and the model reaches optimal when all of them are combined.
\end{abstract}

%%%%%%%%% BODY TEXT
\section{Model Description}

The task of visual relationship detection can be defined as a mapping $f$ from image $I$ to 3 labels and 2 boxes $l_S,l_P,l_O,b_S,b_O$
\begin{equation}
I \xrightarrow{f} l_S,l_P,l_O,b_S,b_O
\end{equation}
where $l,b$ stand for labels and boxes, $S,P,O$ stand for subject, predicate, object. We decompose $f$ into object detector $f_{det}$ and relationship classifier $f_{rel}$:
\begin{equation}
I \xrightarrow{f_{det}} l_S,l_O,b_S,b_O,v_S,v_O \xrightarrow{f_{rel}} l_P
\end{equation}
The decomposition means that we can run an object detector on the input image to obtain labels, boxes and visual features for subject and object, then use these as input features to the relationship classifier which only needs to output a label. There are two obvious advantages in this model: 1) learning complexity is dramatically reduced, since we can simply use an off-the-shelf object detector as $f_{det}$ without the need for re-training, hence the learn-able weights exist only in the small subnet $f_{rel}$; 2) We have much richer features for relationships, \ie, $l_S,l_O,b_S,b_O,v_S,v_O$ for $f_{rel}$, instead of only the image $I$ for $f$.

We further assume that the semantic feature $l_S,l_O$, spatial feature $b_S,b_O$ and visual feature $v_S,v_O$ are independent from each other. So we can build 3 separate branches of sub-networks for them. This is the basic work flow of our model.

\begin{figure*}
  \centering
  \includegraphics[width=\textwidth]{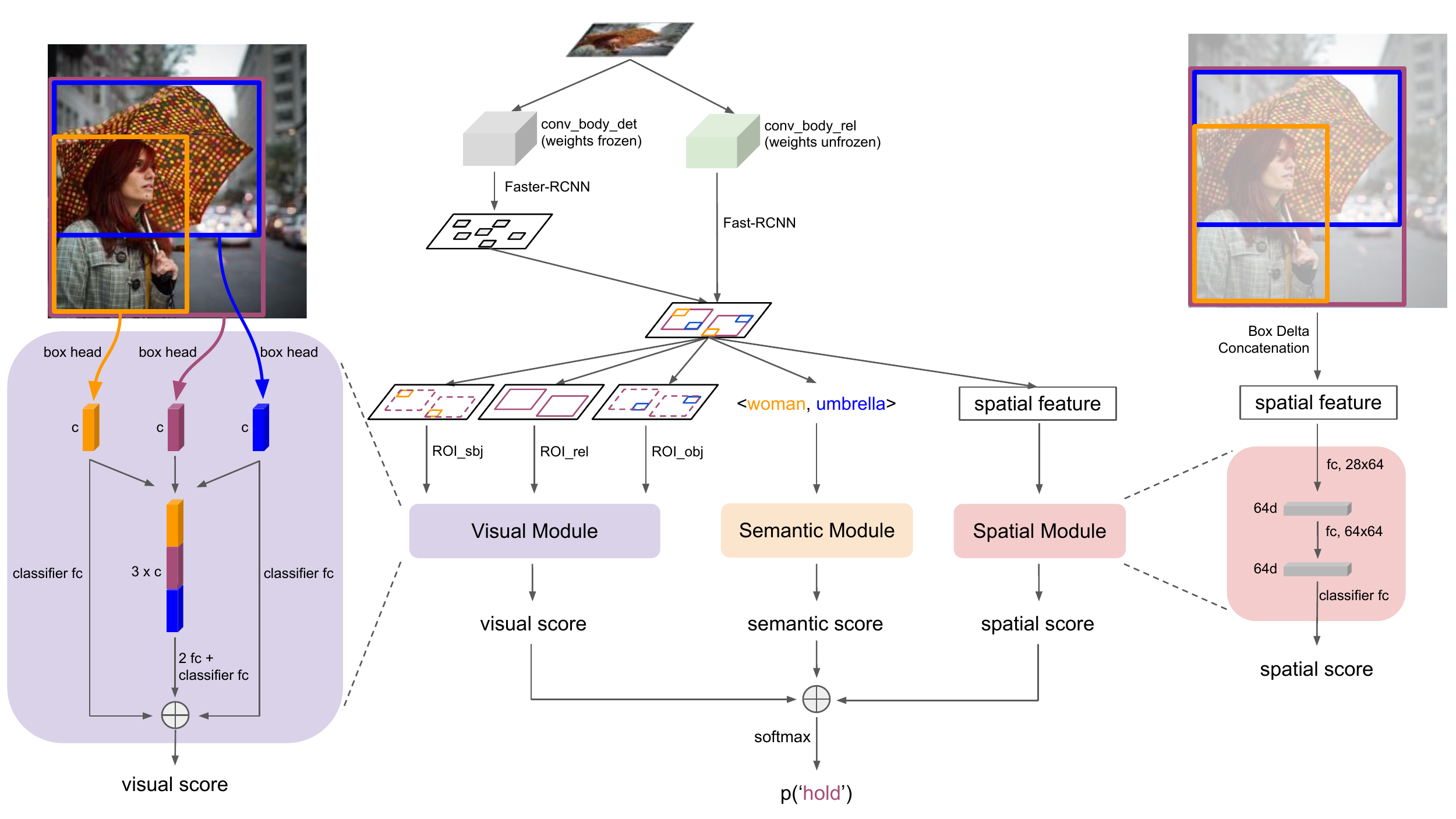}
  \setlength\belowcaptionskip{-2ex}
  \caption{Model Architecture}
  \label{fig:architecture}
\end{figure*}

Figure\ref{fig:architecture} shows our model in details. The network takes an input image and outputs the 6 aforementioned features, then each branch uses its corresponding feature to produce a confidence score for predicates, then all scores are added up and normalized by softmax. We now introduce each module's design and their motivation.

\subsection{Relationship Proposal}

A relationship proposal is defined as a pair of objects that is very likely related\cite{zhang2017relationship}. In our model we first detect all meaningful objects by running an object detector, then we simply consider each pair of objects is a relationship proposal. The following modules learn to classify each pair as either ``no relationship'' or one of the $9$ predicates, not including the ``is'' relationship.

\subsection{Semantic Module}

Zeller, et al.\cite{zellers2018scenegraphs} introduced a frequency baseline that performs reasonably well on Visual Genome dataset\cite{xu2017scene} by counting frequencies of predicates given subject and object. Its motivation is that in general cases, the types of relationships between two objects are usually limited, \eg, given the subject being person and object being horse, their relationship is highly likely to be “ride”, “walk”, “feed”, but less likely to be “stand on”, “carry”, “wear”, \etc. In short, the $\langle subject,predicate,object\rangle$ composition is usually biased. Furthermore, there are numerous types of possible relationships, and any relationship detection dataset can only contain a limited number of them, making the bias even stronger.

We improved this baseline by removing the background class of subject and object and used it as our baseline. Specifically, for each training image we count the occurrence of $l_P$ given $l_S,l_O$ in the ground truth annotations, and we end up with an empirical distribution $p(P|S,O)$ for the whole training set. We do this under the assumption that the test set is also drawn from the same distribution. We then build the remaining modules to learn a complementary residual on top of the output of this baseline.

\subsection{Spatial Module}

In the challenge dataset, the three predicates ``on'', ``under'', ``inside\_of'' indicate purely spatial relationships \ie, the relative locations of subject and object are sufficient to tell the relationship. A common solution, as applied in Faster-RCNN\cite{ren2015faster}, is to learn a mapping from visual features to location offsets. However, the learning becomes significantly hard when the distance of two objects are very far\cite{gkioxari2017interactnet}, which is often the case for relationships. We capture spatial information by encoding the box coordinates of subjects and objects using box delta\cite{ren2015faster} and normalized coordinates:
\begin{equation}
\langle\Delta(b_S,b_O),\Delta(b_S,b_P),\Delta(b_P,b_O),c(b_S),c(b_O)\rangle
\end{equation}
where $\Delta(b_1,b_2)$ are box delta of two boxes $b_1,b_2$, and $c(b)$ are normalized coordinates of box $b$, which are defined as:
\begin{equation}
\Delta(b_1,b_2)=\langle\frac{x_1-x_2}{w_2},\frac{y_1-y_2}{h_2},\log\frac{w_1}{w_2},\log{h_1}{h_2}\rangle
\end{equation}
\begin{equation}
c(b)=\langle\frac{x_{min}}{w},\frac{y_{min}}{h},\frac{x_{max}}{w},\frac{y_{max}}{h},\frac{a_{box}}{a_{img}}\rangle
\end{equation}
where $b_1=(x_1,y_1,w_1,h_1)$ and $b_2=(x_2,y_2,w_2,h_2)$, $w,h$ are width and height of the image, $a_{box}$ and $a_{img}$ are areas of the box and image.

\begin{figure*}[t!]
  \centering
  \begin{subfigure}{1.34\columnwidth}
    \centering
    \begin{subfigure}{0.48\columnwidth}
      \centering
      \includegraphics[width=\textwidth]{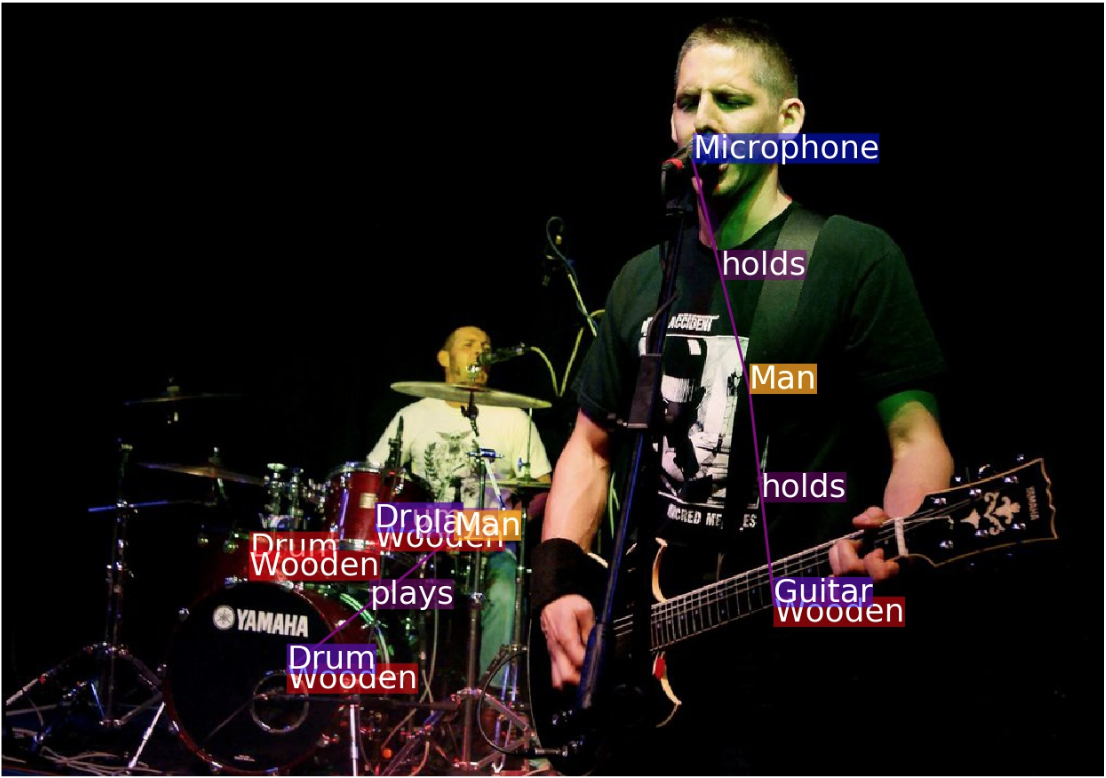}
    \end{subfigure}
    \centering
    \begin{subfigure}{0.48\columnwidth}
      \centering
      \includegraphics[width=\textwidth]{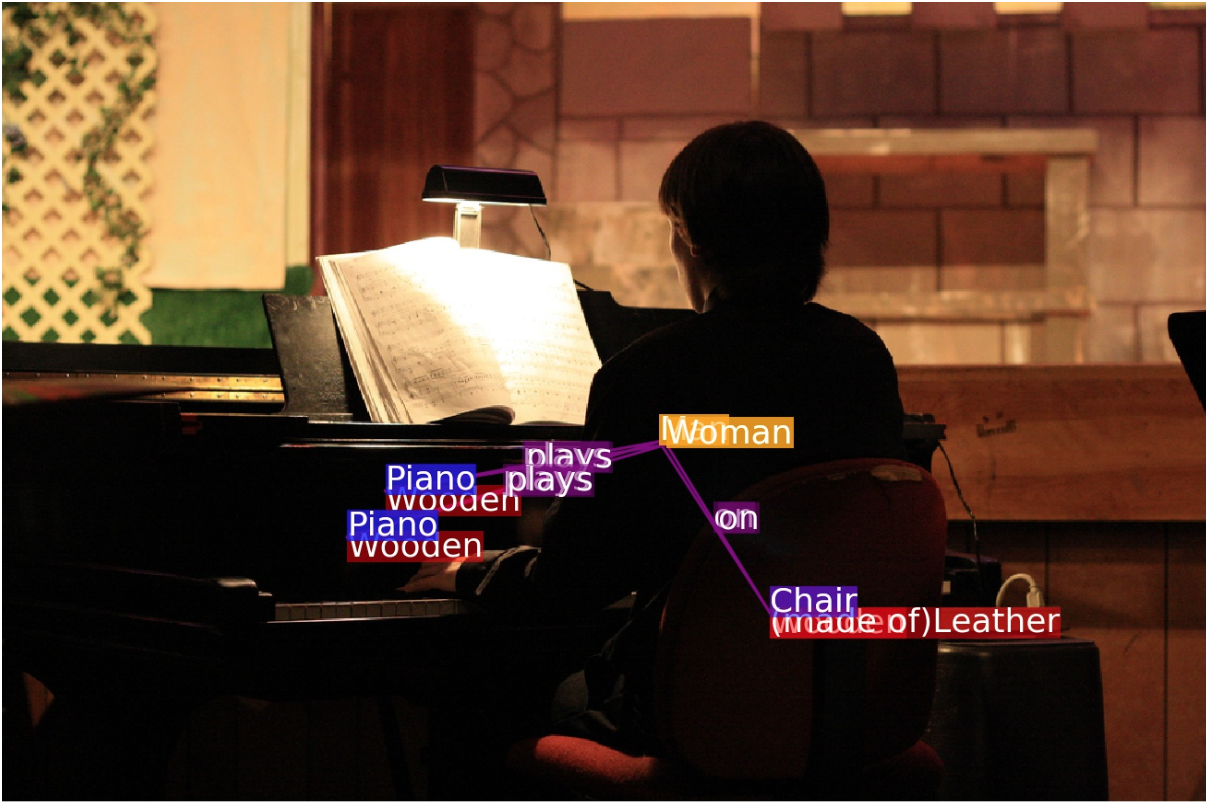}
    \end{subfigure}
    \centering
    \begin{subfigure}{0.48\columnwidth}
      \centering
      \includegraphics[width=\textwidth]{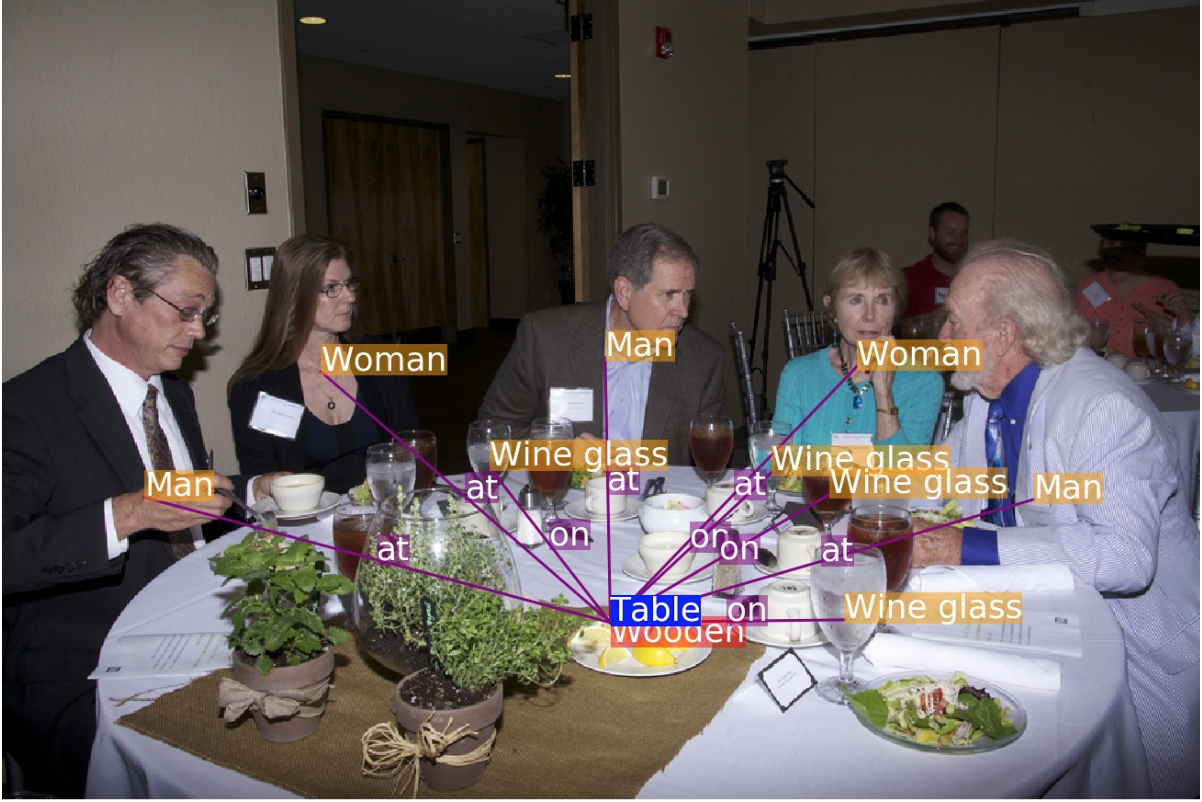}
    \end{subfigure}
    \centering
    \begin{subfigure}{0.48\columnwidth}
      \centering
      \includegraphics[width=\textwidth]{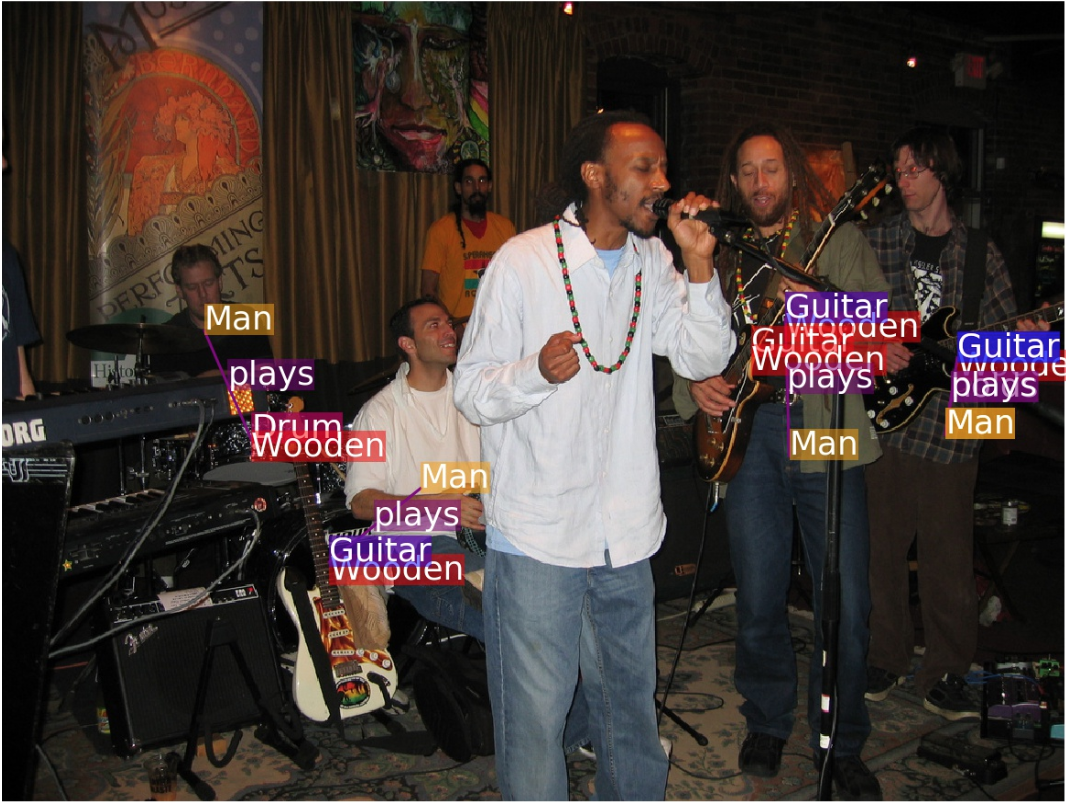}
    \end{subfigure}
  \end{subfigure}
  \centering
  \begin{subfigure}{0.66\columnwidth}
    \centering
    \begin{subfigure}{\columnwidth}
      \centering
      \includegraphics[width=\textwidth]{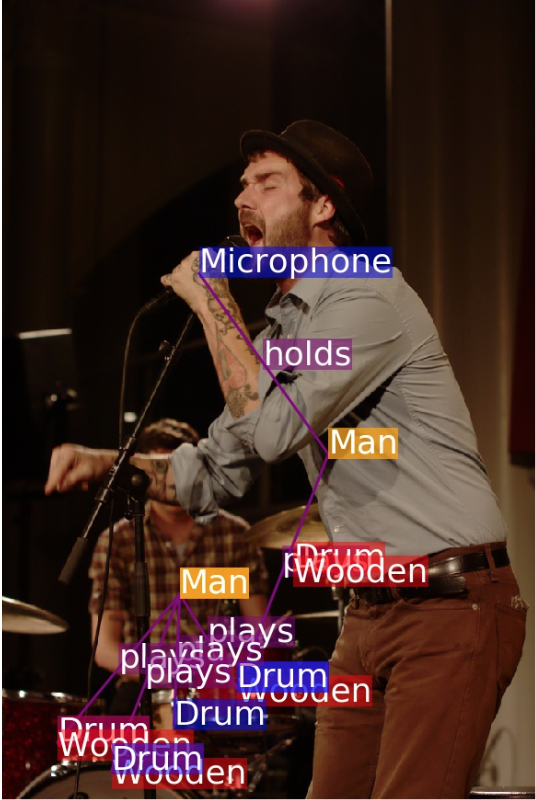}
    \end{subfigure}
  \end{subfigure}
\setlength\belowcaptionskip{-2ex}
\captionsetup{font=small}
\caption{Qualitative results}
\label{fig:qualitative}
\end{figure*}

\subsection{Visual Module}

Visual Module is useful mainly for three reasons: 1) it accounts for all other types of relationships that spatial features can hardly predict, \eg, interactions such as ``man play guitar'' and ``woman wear handbag''; 2) it solves relationship reference problems\cite{krishna2018referring}, \ie, when there are multiple subjects or objects that belong to a same category, we need to know which subject is related to which object; 3) for some specific interactions, \eg, ``throw'', ``eat'' ``ride'', the visual appearance of the subject or object alone is very informative about the predicate. With these motivations, we feed subject, predicate, object ROIs into the backbone and get the feature vectors from its last fc layer as our visual features, then we concatenate these three features and feed them into 2 additional randomly initialized fc layers followed by an extra fc layer to get a logit, \ie, unnormalized score. We also add one fc layer on top of the subject feature and another fc layer on top of the object feature to get two scores. These two scores are the predictions made solely by the subject/object feature according to the third reason mentioned above.

\subsection{The ``is'' Relationship}

In this challenge, ``$\langle object \rangle$ is $\langle attribute \rangle$'' is also considered as relationships, where there is only one object involved. We achieve this sub-task by using a completely separate, single-branch, Fast-RCNN based model. We use the same object detector to get proposals for this model, then for each proposal the model produces a probability distribution over all attributes with the Fast-RCNN pipeline.

\section{Implementation}

% \subsection{Dataset Cleaning}

% We found that for those non-attribute relationships, there are only 57 annotated object classes instead of 61 shown on the website, so we keep only the annotations of those 57 objects. For those attribute relationships, there are only 23 object classes which is a subset of the 57 classes, thus we directly use the same object detector but exclude objects that are not among the 23 classes. We split training images into attribute and non-attribute ones, but we keep only one validation set which is officially suggested.

\subsection{Training}

We train the regular relationship model and the ``is'' model separately. For the former, we train for 8 epochs using the default hyper-parameter settings from Detectron\cite{Detectron2018}. We copy layers from the first conv layer to the last fc layer as the relationship feature extractor. We freeze the the object detector's weights but set the relationship branch learn-able. We also tried setting object detector free for fine-tuning, and found that it over-fits to the few objects that appear in relationship annotations and loses generalization ability during testing. We set the ratio of negative and positive as 3 during ROI sampling. For the latter, we train for 8 epochs as well using the same default settings. We set the negative and positive ratio as 1 as we found it optimal.

\subsection{Testing}

For non-attribute relationships, we obtain a relationship score by
\begin{equation}
S_{SPO} = S_S \times S_P \times S_O
\end{equation}
where $S_S$ and $S_O$ are obtained from object detector, $S_P$ is the output of our relationship model.
For attribute relationships, we obtain the score by
\begin{equation}
S_{OA} = S_O \times S_A
\end{equation}
where $S_O$ and $S_A$ are scores of objects and attributes. We use $S_{SPO}$ and $S_{OA}$ to rank all predictions and get the top 200 as final results.

\section{Experiments}

\begin{table}[t]
\centering
\resizebox{\columnwidth}{!}{
\begin{tabular}{c c c c c c}
\hline
 & R@50 & mAP\_rel & mAP\_phr & Score on val & Score on public \\
 Baseline & 72.98 & 26.54 & 32.77 & 38.32 & 22.21 (2nd) \\
 $\langle S,P,O \rangle$ & 74.13 & 32.41 & 39.55 & 43.61 & -\\
 $\langle S,P,O \rangle + S + O$ & 74.46 & 34.16 & 39.59 & 44.39 & -\\
 $\langle S,P,O \rangle + S + O + spt$ & 74.40 & 34.96 & 40.70 & 45.14 & 33.21 (1st)\\
\hline
\end{tabular}
}
\caption{Ablation Study}
\label{tab:abl}
\end{table}

\subsection{Ablation Study}

We show performance of four models with the following settings: 1) \textbf{baseline}: only the semantic module. 2)$\langle \textbf{S,P,O} \rangle$: using semantic module and visual module without the direct predictions from subject/object. 3) $\langle \textbf{S,P,O} \rangle \textbf{+ S + O}$: using semantic module and the complete visual module 4) $\langle \textbf{S,P,O} \rangle \textbf{+ S + O + spt}$: our complete model.

\subsection{Qualitative Results}

We show several example outputs of our model. We can see from Figure\ref{fig:qualitative} that we are able to correctly refer relationships, \ie, when there are multiple people playing multiple guitars, our model accurately points to the truly related pairs. One major failure case of our model is on the predicate ``hold'', where the model usually needs to focus on the small area of the intersection of a human hand and the object, which our model is currently not expert at.

{\small
\bibliographystyle{ieee}
\bibliography{egbib}
}

\clearpage

\end{document}